\documentclass[runningheads]{llncs}
\usepackage[T1]{fontenc}
\usepackage{graphicx}
\usepackage{cite}
\usepackage{amsfonts}
\usepackage[ruled]{algorithm2e}
\usepackage{verbatim}
\usepackage{textcomp}
\usepackage{subcaption}
\usepackage{threeparttable}
\usepackage{url}
\usepackage[table,xcdraw]{xcolor}
\usepackage{colortbl}
\usepackage{booktabs}
\usepackage{wrapfig}
\usepackage{float}
\usepackage{multirow}
\usepackage{enumitem}
\usepackage{makecell}
\usepackage{setspace}
\usepackage{orcidlink}

\usepackage{cite}
\usepackage{amsmath,amssymb,amsfonts}

\newcommand{\eg}{{\em e.g.}}           % e.g.
\newcommand{\ie}{{\em i.e.}}           % i.e.
\begin{document}
\title{3D Medical Image Segmentation with Sparse Annotation via Cross-Teaching between 3D and 2D Networks}
\titlerunning{3D-2D Cross-Teaching}

\author{
%1{Cai, Heng}2{Qi, Lei}3{Yu, Qian}4{Shi, Yinghuan}5{Gao, Yang}
Heng Cai\inst{1} \and
Lei Qi\inst{2} \and
Qian Yu\inst{3} \and Yinghuan Shi\inst{1,}\thanks{Corresponding author.} \and Yang Gao\inst{1}}

\authorrunning{H. Cai et al.}

\institute{State Key Laboratory of Novel Software Technology, \\National Insititute of Health-care Data Science, Nanjing University, China \\ \email{echo@smail.nju.edu.cn}, \email{\{syh,gaoy\}@nju.edu.cn} \and
School of Computer Science and Engineering, Southeast University, China\\
\email{qilei@seu.edu.cn}\\
 \and
School of Data and Computer Science, Shandong Woman's University, China\\
\email{yuqian@sdwu.edu.cn}}

\maketitle              % typeset the header of the contribution
\begin{abstract}
Medical image segmentation typically necessitates a large and precisely annotated dataset. However, obtaining pixel-wise annotation is a labor-intensive task that requires significant effort from domain experts, making it challenging to obtain in practical clinical scenarios. In such situations, reducing the amount of annotation required is a more practical approach. One feasible direction is sparse annotation, which involves annotating only a few slices, and has several advantages over traditional weak annotation methods such as bounding boxes and scribbles, as it preserves exact boundaries. However, learning from sparse annotation is challenging due to the scarcity of supervision signals. To address this issue, we propose a framework that can robustly learn from sparse annotation using the cross-teaching of both 3D and 2D networks. Considering the characteristic of these networks, we develop two pseudo label selection strategies, which are hard-soft confidence threshold and consistent label fusion. Our experimental results on the MMWHS dataset demonstrate that our method outperforms the state-of-the-art (SOTA) semi-supervised segmentation methods. Moreover, our approach achieves results that are comparable to the fully-supervised upper bound result. Our code is available at \textcolor{magenta}{\url{https://github.com/HengCai-NJU/3D2DCT}}.

\keywords{ 3D segmentation  \and Sparse annotation \and Cross-teaching}
\end{abstract}
\section{Introduction}
Medical image segmentation is greatly helpful to diagnosis and auxiliary treatment of diseases. Recently, deep learning methods~\cite{milletari2016v,hatamizadeh2022unetr} has largely improved the performance of segmentation. However, the success of deep learning methods typically relies on large densely annotated datasets, which require great efforts from domain experts and thus are hard to obtain in clinical applications.

To this end, many weakly-supervised segmentation (WSS) methods are developed to alleviate the annotation burden, including image level~\cite{lee2019ficklenet,lee2021anti}, bounding box~\cite{song2019box,kulharia2020box2seg,oh2021background}, scribble~\cite{lin2016scribblesup,xu2021scribble} and even points~\cite{bearman2016s,li2022fully}. These methods utilize weak label as supervision signal to train the model and produce segmentation results. Unfortunately, the performance gap between these methods and its corresponding upper bound (\ie, the result of fully-supervised methods) is still large. The main reason is that these annotation methods do not provide the information of object boundaries, which are crucial for segmentation task.

A new annotation strategy has been proposed and investigated recently. It is typically referred as sparse annotation and it only requires a few slice of each volume to be labeled. With this annotation way, the exact boundaries of different classes are precisely kept. It shows great potential in reducing the amount of annotation. And its advantage over traditional weak annotations has been validated in previous work~\cite{bitarafan20203d}. To enlarge the slice difference, we annotate slices from two different planes instead of from a single plane. 

Most existing methods solve the problem by generating pseudo label through registration.~\cite{bitarafan20203d} trains the segmentation model through an iterative step between propagating pseudo label and updating segmentation model.~\cite{li2022pln} adopts mean-teacher framework as segmentation model and utilizes registration module to produce pseudo label.~\cite{cai2023orthogonal} proposes a co-training framework to leverage the dense pseudo label and sparse orthogonal annotation. Though achieving remarkable results, the limitation of these methods cannot be ignored. These methods rely heavily on the quality of registration result. When the registration suffers due to many reasons (\eg, small and intricate objects, large variance between adjacent slices), the performance of segmentation models will be largely degraded.

Thus, we suggest to view this problem from the perspective of semi-supervised segmentation (SSS). Traditional setting of 3D SSS is that there are several volumes with dense annotation and a large number of volumes without any annotation. And now there are voxels with annotation and voxels without annotation in every volume. This actually complies with the idea of SSS, as long as we view the labeled and unlabeled voxels as labeled and unlabeled samples, respectively. 

SSS methods~\cite{chen2021semi,shi2021inconsistency,yang2023revisiting} mostly fall into two categories, 1) entropy minimization and 2) consistency regularization. And one of the most popular paradigms is co-training~\cite{zhou2019semi,xia2020uncertainty,chen2021semi}. 
%\cite{xia2020uncertainty} trains 3D networks on different views and enforces consistent prediction of them. \cite{zhou2019semi} utilizes the prediction of 2D models trained on different planes to produce more accurate pseudo label. \cite{chen2021semi} imposes consistency on the prediction of two networks for the same input image and uses one-hot prediction as pseudo label.
Inspired by these co-training methods, we propose our method based on the idea of cross-teaching. As co-training theory conveys, the success of co-training largely lies on the view-difference of different networks~\cite{qiao2018deep}. Some works encourage the difference by applying different transformation to each network. \cite{luo2022semi} directly uses two type of networks (\ie, CNN and transformer) to guarantee the difference. Here we further extend it by adopting networks of different dimensions (\ie, 2D CNN and 3D CNN). 3D network and 2D network work largely differently for 3D network involves the inter-slice information while 2D network only utilize inner-slice information. The 3D network is trained on volume with sparse annotation and we use two 2D networks to learn from slices of two different planes. Thus, the view difference can be well-preserved.

However, it is still hard to directly train with the sparse annotation due to limited supervision signal. So we utilize 3D and 2D networks to produce pseudo label to each other. In order to select more credible pseudo label, we specially propose two strategies for the pseudo label selection of 3D network and 2D networks, respectively. For 3D network, simply setting a prediction probability threshold can exclude those voxels with less confidence, which are more likely to be false prediction. However, Some predictions with high quality but low confidence are also excluded. Thus, we estimate the quality of each prediction, and design hard-soft thresholds. If the prediction is of high quality, the voxels that overpass the soft threshold are selected as pseudo label. Otherwise, only the voxels overpassing the hard threshold can be used to supervise 2D networks. For 2D networks, compared with calculating uncertainty which introduces extra computation cost, we simply use the consistent prediction of two 2D networks. As the two networks are trained on slices of different planes, thus their consistent predictions are very likely to be correct. We validate our method on the MMWHS~\cite{zhuang2016multi,zhuang2018multivariate} dataset, and the results show that our method is superior to SOTA semi-supervised segmentation methods in solving sparse annotation problem. Also, our method only uses 16\% of labeled slices but achieves comparable results to the fully supervised method.

To sum up, our contributions are three folds:
\begin{itemize}
    \item A new perspective of solving sparse annotation problem, which is more versatile compared with recent methods using registration.
    \item A novel cross-teaching paradigm which imposes consistency on the prediction of 3D and 2D networks. Our method enlarges the view difference of networks and boosts the performance.
    \item A pseudo label selection strategy discriminating between reliable and unreliable predictions, which excludes error-prone voxels while keeping credible voxels though with low confidence.
\end{itemize}

\begin{figure}[t!]
\centering
\begin{subfigure}[b]{0.15\textwidth}
\centering
\includegraphics[width=\textwidth]{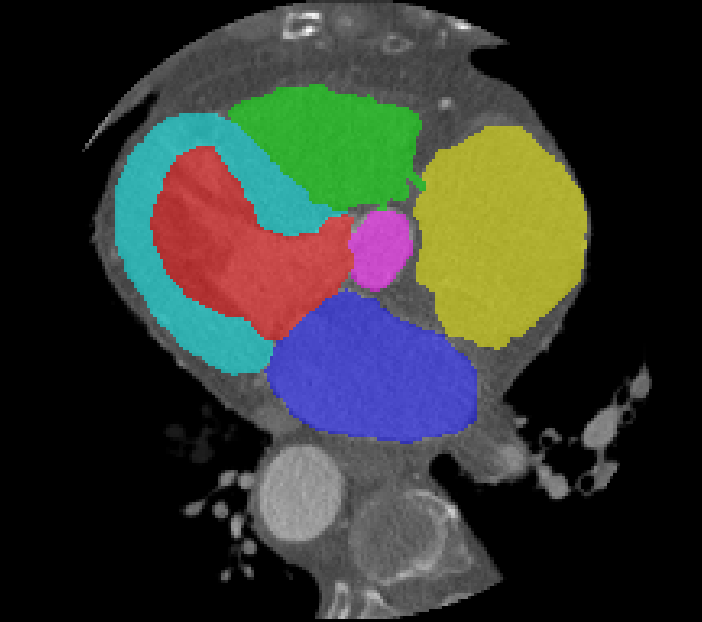}
\caption{}
\end{subfigure}
\begin{subfigure}[b]{0.15\textwidth}
\centering
\includegraphics[width=\textwidth]{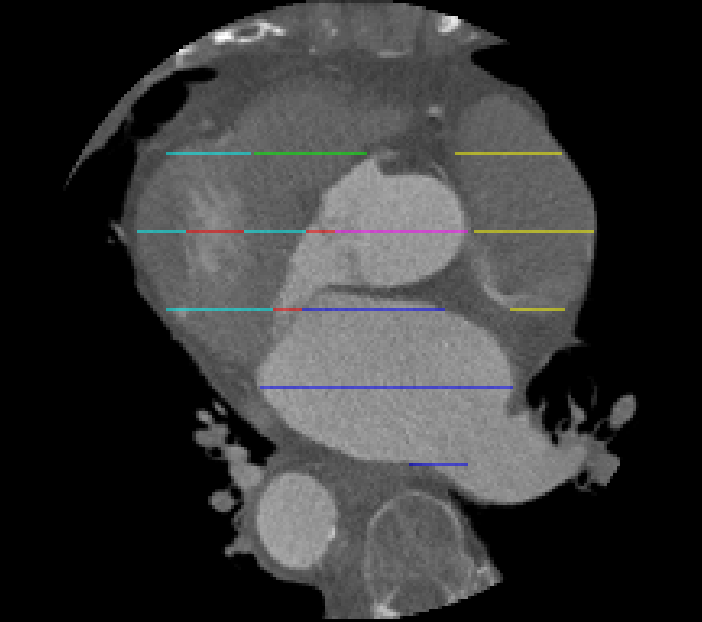}
\caption{}
\end{subfigure}
\begin{subfigure}[b]{0.15\textwidth}
\centering
\includegraphics[width=\textwidth]{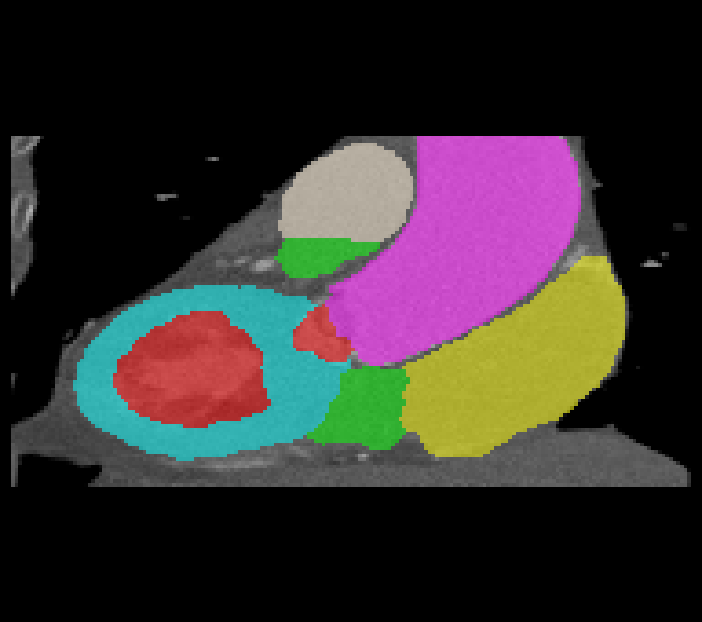}
\caption{}
\end{subfigure}
\begin{subfigure}[b]{0.15\textwidth}
\centering
\includegraphics[width=\textwidth]{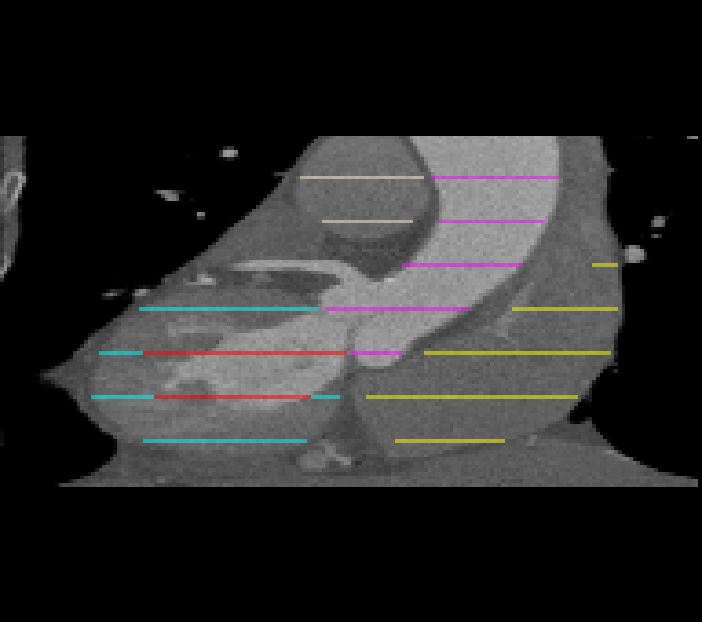}
\caption{}
\end{subfigure}
\begin{subfigure}[b]{0.15\textwidth}
\centering
\includegraphics[width=\textwidth]{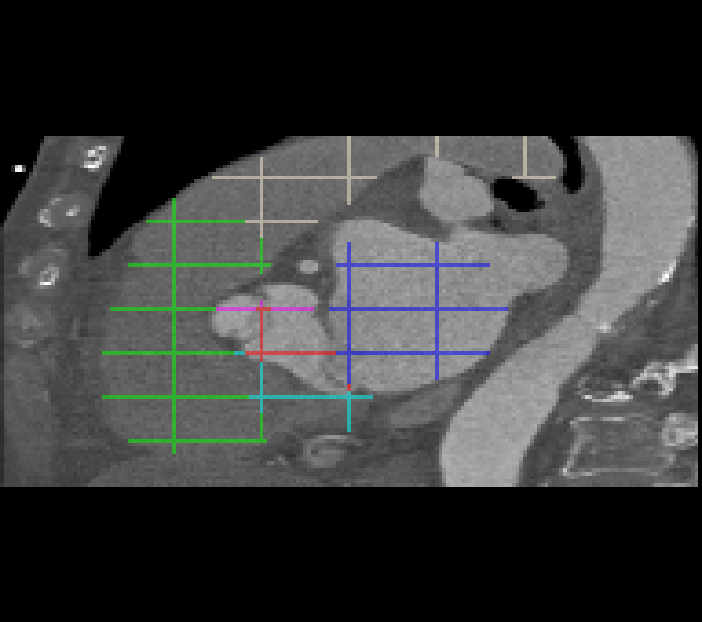}
\caption{}
\end{subfigure}
\begin{subfigure}[b]{0.15\textwidth}
\centering
\includegraphics[width=\textwidth]{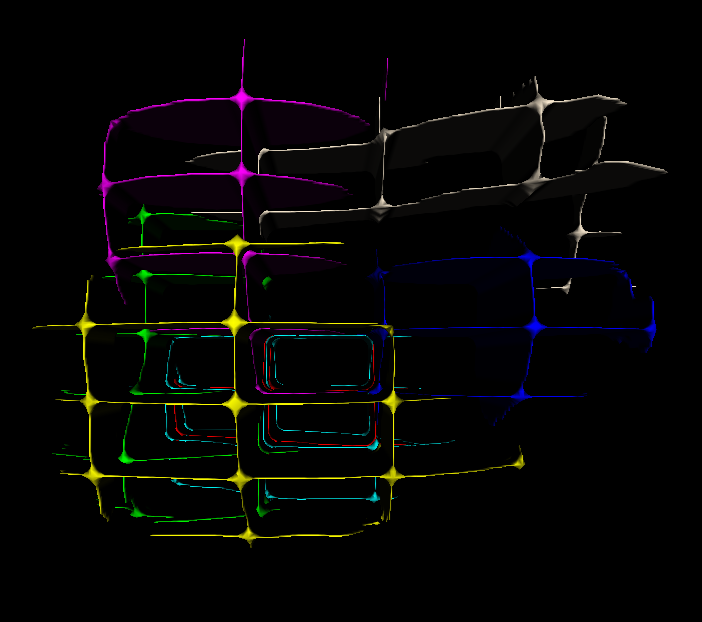}
\caption{}
\end{subfigure}

\caption{An example of cross annotaion. (a) and (b) are typical annotations of transverse plane slices; (c) and (d) are typical annotations of coronal plane slices; (e) is a typical annotation of saggital plane slices and (f) is the 3D view of cross annotation.}
\label{fig:annotation}
\end{figure}

\section{Method}
\subsection{Cross Annotation}
Recent sparse annotation methods \cite{bitarafan20203d,li2022pln} only label one slice for each volume, however, this annotation has many limitations. 1) The segmentation object must be visible on the labeled slice. Unfortunately, in most cases, the segmentation classes cannot be all visible in a single slice, especially in multi-class segmentation tasks. 2) Even though there is only one class and is visible in the labeled slice, the variance between slices might be large, and thus the information provided by a single slice is not enough to train a well-performed segmentation model. Based on these two observations, we label multiple slices for each volume. Empirically, the selection of slices should follow the rule that they should be as variant as possible in order to provide more information and have a broader coverage of the whole data distribution. Thus, we label slices from two planes (\eg, transverse plane and coronal plane) because the difference involved by planes is larger than that involved by slice position on a single plane.
\begin{figure}[t]
    \centering
    \includegraphics[width = 1\textwidth]{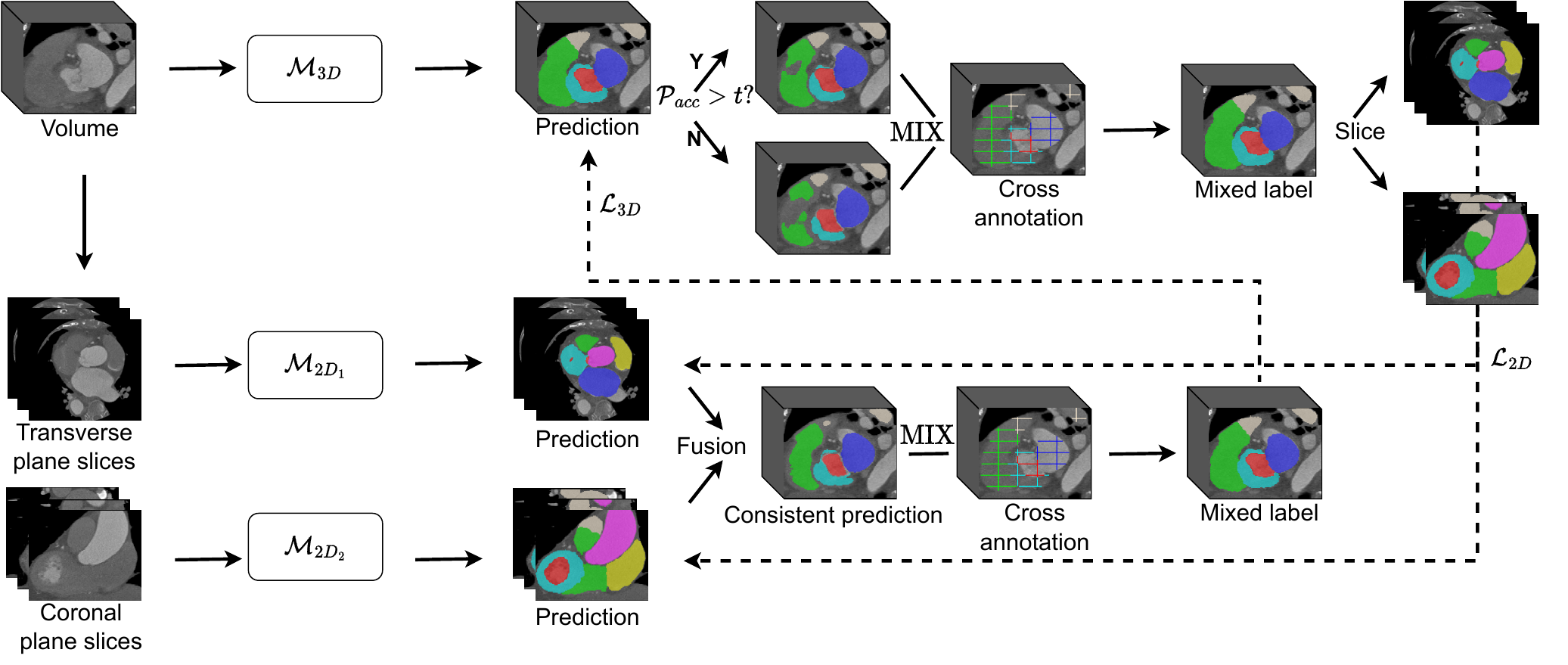}
    \caption{Overview of the proposed 3D-2D cross-teaching framework. For a volume with cross annotation, 3D network and 2D networks give predictions of it. We use hard-soft threshold and consistent prediction fusion to select credible pseudo label. Then the pseudo label is mixed with ground truth sparse annotation to supervise other networks. }
    \label{fig:framework}
\end{figure}
The annotation looks like crosses from the third plane, so we name it $\bold{Cross~ Annotation}$. The illustration of cross annotation is shown in Fig.~\ref{fig:annotation}. Furthermore, in order to make the slices as variant as possible, we simply select those slices with a same distance. And the distance is set according to the dataset. For example, the distance can be large for easy segmentation task with lots of volumes. Otherwise, the distance should be closer for difficult task or with less volumes. And here we provide a simple strategy to determine the distance.

First label one slice for each plane in a volume, and train the model to monitor its performance on validation set, which has ground truth dense annotation. Then halve the distance (\ie, double the labeling slice), and test the trained model on validation set again. The performance gain can be calculated. Then repeat the procedure until the performance gain is less than half of the previous gain. The current distance is the final distance. The performance gain is low by labeling more slices.

The aim of the task is to train a segmentation model on dataset $\mathcal{D}$ consists of $L$ volumes $X_1,X_2,...,X_L$ with cross annotation $Y_1,Y_2,...,Y_L$. 
\subsection{3D-2D Cross Teaching}
Our framework consists of three networks, which are a 3D networks and two 2D networks. We leverage the unlabeled voxels through the cross teaching between 3D network and 2D networks. Specifically, the 3D network is trained on volumes and the 2D networks are trained with slices on transverse plane and coronal plane, respectively. The difference between 3D and 2D network is inherently in their network structure, and the difference between 2D networks comes from the different plane slices used to train the networks. 

For each sample, 3D network directly use it as input. Then it is cut into slices from two directions, resulting in transverse and coronal plane slices, which are used to train the 2D networks. And the prediction of each network, which is denoted as $P$, is used as pseudo label for the other network after selection. The selection strategy is detailedly introduced in the following part. 

To increase supervision signal for each training sample, we mix the selected pseudo label and ground truth sparse annotation together for supervision. And it is formulated as:
\begin{equation}
\hat Y=\textrm{MIX}(Y,P),
\end{equation}
where $\textrm{MIX}(\cdot,\cdot)$ is a function that replaces the label in $P$ with the label in $Y$ for those voxels with ground truth annotation. 

Considering that the performance of 3D network is typically superior to 2D networks, we further introduce a label correction strategy. If the prediction of 3D network and the pseudo label from 2D networks differ, no loss on that particular voxel should be calculated as long as the confidence of 3D networks is higher than both 2D networks. We use $M$ to indicate how much a voxel contribute to the loss calculation, and the value of position $i$ is 0 if the loss of voxel $i$ should not be calculated, otherwise 1 for ground truth annotation and $w$ for pseudo label, where $w$ is a value increasing from 0 to 0.1 according to ramp-up from~\cite{Laine2016Temporal}.

The total loss consists of cross-entropy loss and dice loss:
\begin{equation}
\mathcal{L}_{ce}=-\frac{1}{\sum_{i=1}^{H\times W\times D}m_i}\sum_{i=1}^{H\times W\times D}m_iy_i\log p_i,
\end{equation}
and
\begin{equation}\label{(2)}
\mathcal{L}_{dice}=1-\frac{2\times\sum_{i=1}^{H\times W\times D}m_ip_iy_i}{\sum_{i=1}^{H\times W\times D}m_i (p_i^2+ y_i^2)},
\end{equation}
where $p_i$, $y_i$ is the output and the label in $\hat Y$ of voxel $i$, respectively. $m_i$ is the value of $M$ at position $i$. $H, W, D$ denote the height, width and depth of the input volume, respectively. 
And the total loss is denoted as:
\begin{equation}\label{(3)}
\mathcal{L}=\frac{1}{2}\mathcal{L}_{ce}+\frac{1}{2}\mathcal{L}_{dice}.
\end{equation}
\subsection{Pseudo Label Selection}
\subsubsection{Hard-Soft Confidence Threshold.}
Due to the limitation of supervision signal, the prediction of 3D model has lots of noisy label. If it is directly used as pseudo label for 2D networks, it will cause a performance degradation on 2D networks. So we set a confidence threshold to select voxels which are more likely to be correct. However, we find that this may also filter out correct prediction with lower confidence, which causes the waste of useful information. If we know the quality of the prediction, we can set a lower confidence threshold for the voxels in the prediction of high-quality in order to utilize more voxels. However, the real accuracy $R_{acc}$ of prediction is unknown, for the dense annotation is unavailable during training. What we can obtain is the pseudo accuracy $P_{acc}$ calculated with the prediction and the sparse annotation. And we find that $R_{acc}$ and $P_{acc}$ are completely related on the training samples. Thus, it is reasonable to estimate $R_{acc}$ using $P_{acc}$:

\begin{equation}
    R_{acc}\approx P_{acc}=\sum_{i=1}^{H\times W\times D}\mathbb{I}(\hat p_i=y_i)/(H\times W\times D),
\end{equation}
where $\mathbb{I}(\cdot)$ is the indicator function and $\hat p_i$ is the one-hot prediction of voxel $i$.

Now we introduce our hard-soft confidence threshold strategy to select from 3D prediction. We divide all prediction into reliable prediction (\ie, with higher $P_{acc}$) and unreliable prediction (\ie, with lower $P_{acc}$) according to threshold $t_q$. And we set different confidence thresholds for these two types of prediction, which are soft threshold $t_s$ with lower value and hard threshold $t_h$ with higher value. In reliable prediction, voxels with confidence higher than soft threshold can be selected as pseudo label. The soft threshold aims to keep the less confident voxels in reliable prediction and filter out those extremely uncertain voxels to reduce the influence of false supervision. And in unreliable prediction, only those voxels with confidence higher than hard threshold can be selected as pseudo label. The hard threshold is set to choose high-quality voxels from unreliable prediction. The hard-soft confidence threshold strategy achieves a balance between increasing supervision signals and reducing label noise.
\subsubsection{Consistent Prediction Fusion.}
 Considering that 2D networks are not able to utilize inter-slice information, their performance is typically inferior to that of 3D network. Simply setting threshold or calculating uncertainty is either of limited use or involving large extra calculation cost. To this end, we provide a selection strategy which is useful and introduces no additional calculation. The 2D networks are trained on slices from different planes and they learn different patterns to distinguish foreground from background. So they will produce predictions with large diversity for a same input sample and the consensus of the two networks are quite possible to be correct. Thus, we use the consistent part of prediction from the two networks as pseudo label for 3D network.

\begin{table}[t]
\caption{Comparison result on MMWHS dataset.}\label{tab:sotacompare}
\centering
\begin{threeparttable}
\resizebox{1\linewidth}{!}{
\begin{tabular}{cc|c|c|cccc}
\noalign{\smallskip}
\hline
\noalign{\smallskip}
\multicolumn{2}{c|}{\multirow{2}{*}{Method}} &{\multirow{2}{*}{Venue}}                        & \multirow{1}{*}{Labeled slices}                           & \multicolumn{4}{c}{Metrics}           \\ 
 \cmidrule(r){5-8}
\multicolumn{2}{c|}{}         & {} &{/ volume}   & Dice (\%)$\uparrow$     & Jaccard(\%)$\uparrow$   & HD (voxel)$\downarrow$   & ASD (voxel)$\downarrow$   \\ 
\noalign{\smallskip}
\hline
\noalign{\smallskip}
\multicolumn{1}{c|}{\multirow{5}{*}{Semi-supervised}} & 
\multirow{1}{*}{MT~\cite{tarvainen2017mean}}  &\multirow{1}{*}{NIPS'17} & \multicolumn{1}{c|}{16}
 &76.25$\pm$4.63&64.89$\pm$4.59&19.40$\pm$9.63&5.65$\pm$2.28\\
\multicolumn{1}{c|}{}& UAMT~\cite{yu2019uncertainty} &MICCAI'19 & \multicolumn{1}{c|}{16} 
&72.19$\pm$11.69&61.54$\pm$11.46&18.39$\pm$8.16&4.97$\pm$2.30\\

\multicolumn{1}{c|}{}& \multirow{1}{*}{CPS~\cite{chen2021semi}}  &\multirow{1}{*}{CVPR'21} & \multicolumn{1}{c|}{16}
&77.19$\pm$7.04&66.96$\pm$5.90&13.10$\pm$3.60&4.00$\pm$2.06\\
\multicolumn{1}{c|}{}&CTBCT~\cite{luo2022semi}& {MIDL'22}&\multicolumn{1}{c|}{16}
&74.20$\pm$6.04&63.04$\pm$5.36&17.91$\pm$3.96&5.15$\pm$1.38\\

\multicolumn{1}{c|}{}& Ours &this paper & \multicolumn{1}{c|}{16} 
&\textbf{82.67$\pm$4.99}&\textbf{72.71$\pm$5.51}&\textbf{12.81$\pm$0.74}&\textbf{3.72$\pm$0.43}\\
\noalign{\smallskip}
\hline
\noalign{\smallskip}
\multicolumn{1}{c|}{{\color[HTML]{656565}Fully-supervised}}                    & {\color[HTML]{656565}V-Net~\cite{milletari2016v}} & {\color[HTML]{656565}3DV'16}& \multicolumn{1}{c|}{{\color[HTML]{656565}96}}& {\color[HTML]{656565}81.69$\pm$4.93}          & {\color[HTML]{656565}71.36$\pm$6.40}          & {\color[HTML]{656565}16.15$\pm$3.13}          & {\color[HTML]{656565}5.01$\pm$1.29}          \\ 
\noalign{\smallskip} 
\hline
\noalign{\smallskip}

\end{tabular}}
\end{threeparttable}
\end{table}

\begin{figure}[t]
    \centering
    \includegraphics[width = 0.9\textwidth]{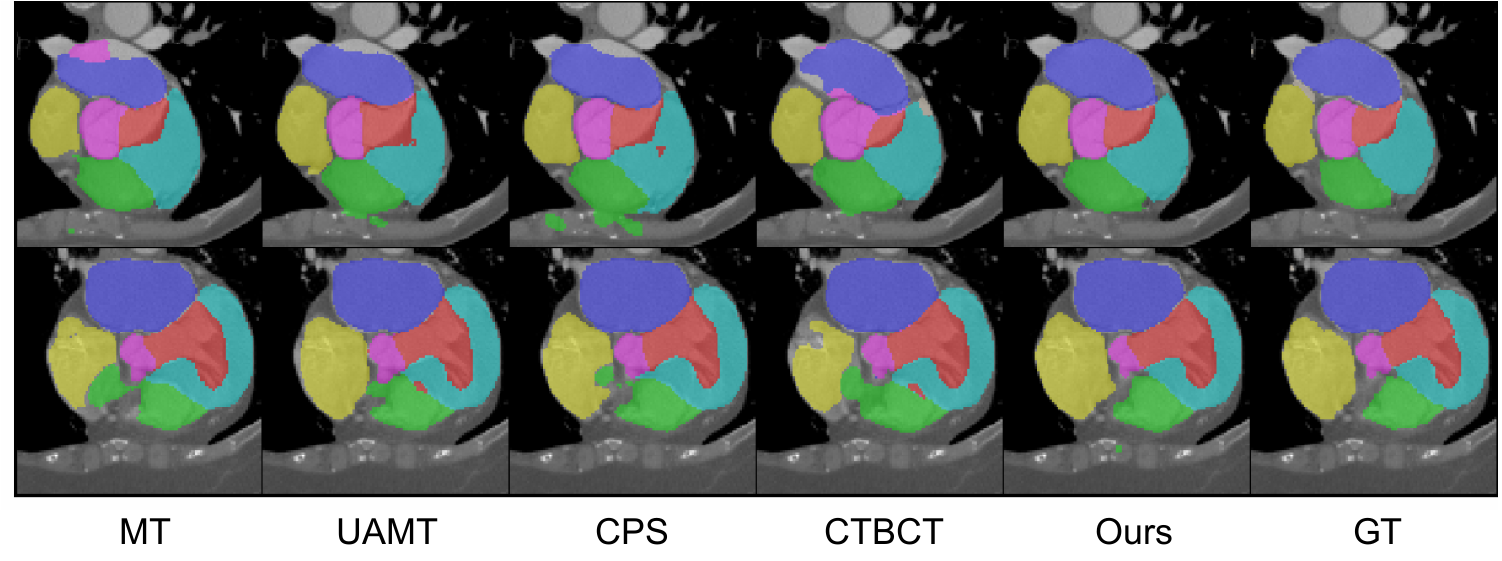}
    \caption{Visual examples of segmentation results on MMWHS dataset.}
    \label{fig:visual}
\end{figure}

\section{Experiments}
\subsection{Dataset and Implementation Details}
\subsubsection{MMWHS Dataset} \cite{zhuang2016multi,zhuang2018multivariate}~is from the MICCAI 2017 challenge, which consists of 20 cardiac CT images with publicly accessible annotations that cover seven whole heart substructures. We split the 20 volumes into 12 for training, 4 for validation and 4 for testing. And we normalize all volumes through z-score normalization. All volumes are reshaped to [192, 192, 96] with linear interpolation.
\subsubsection{Implementation Details.}
 We adopt Adam~\cite{kingma2014adam} with a base learning rate of 0.001 as optimizer and the weight decay is 0.0001. Batch size is 1 and training iteration is 6000. We adopt random crop as data augmentation strategy and the patch size is [176, 176, 96]. And the hyper-parameters are $t_q=0.98$, $t_h=0.9$, $t_s=0.7$ according to experiments on validation set. For 3D and 2D networks, we use V-Net~\cite{milletari2016v} and U-Net~\cite{ronneberger2015u} as backbone, respectively. All experiments are conducted using PyTorch and 3 NVIDIA GeForce RTX 3090 GPUs.

\subsection{Comparison with SOTA methods}
As previous sparse annotation works~\cite{bitarafan20203d,li2022pln} cannot leverage sparse annotation where there are more than one labeled slice in a volume, to verify the effectiveness of our method, we compare it with SOTA semi-supervised segmentation methods, including Mean Teacher (\textbf{MT})~\cite{tarvainen2017mean}, Uncertainty-aware Mean-Teacher (\textbf{UAMT})~\cite{yu2019uncertainty}, Cross-Pseudo Supervision (\textbf{CPS})~\cite{chen2021semi} and Cross Teaching Between CNN and Transformer (\textbf{CTBCT})~\cite{luo2022semi}. The transformer network in CTBCT is implemented as UNETR~\cite{hatamizadeh2022unetr}. Our method uses the prediction of 3D network as result. For fairer comparisons, all experiments are implemented in 3D manners with the same setting. For the evaluation and comparisons of our method and other methods, we use Dice coefficient, Jaccard coefficient, 95\% Hausdorff Distance (HD) and Average Surface Distance (ASD) as quantitative evaluation metrics. The results are required through three runs with different random dataset split and they are reported as mean value $\pm$ standard deviation. The quantitative results and qualitative results are shown in Table~\ref{tab:sotacompare} and Fig.~\ref{fig:visual}.

\subsection{Ablation Study}
We also investigate how hyper-parameters $t_q$, $t_h$ and $t_s$ affect the performance of the method. We conduct quantitative ablation study on the validation set. The results are shown in Table~\ref{tab:ablation}. Bold font presents best results and underline presents the second best.
\begin{table}[t]
\caption{Ablation study on hyper-parameters.}\label{tab:ablation}
\centering
\begin{threeparttable}
\resizebox{1\linewidth}{!}{
\begin{tabular}{c|c|c|c|cccc}
\noalign{\smallskip}
\hline
\noalign{\smallskip}
\multirow{2}{*}{Parameters}&\multirow{2}{*}{$t_q$}&\multirow{2}{*}{$t_h$}                        & \multirow{2}{*}{$t_s$}   & \multicolumn{4}{c}{Metrics}           \\ 
 \cmidrule(r){5-8}
{set}&{}&{}         & {}   & Dice (\%)$\uparrow$     & Jaccard(\%)$\uparrow$   & HD (voxel)$\downarrow$   & ASD (voxel)$\downarrow$   \\ 
\noalign{\smallskip}
\hline
\noalign{\smallskip}
1&$\qquad 0.98\qquad$ &$\qquad 0.90\qquad$  &$\qquad 0.85\qquad$&82.25$\pm$9.00&72.76$\pm$10.73&9.66$\pm$5.22&3.08$\pm$1.69\\
2&0.98 &0.90  & 0.70&\textbf{83.64$\pm$9.39}&\textbf{74.63$\pm$11.10}&\underline{8.60$\pm$4.75}&\underline{2.77$\pm$1.71}\\
3&0.98 &0.90  & 0.50&\underline{82.60$\pm$9.50}&\underline{73.40$\pm$11.21}&10.74$\pm$6.03&3.40$\pm$1.95\\
\noalign{\smallskip} 
\hline
\noalign{\smallskip}
4&0.95 &0.90  & 0.85&81.08$\pm$9.93&71.98$\pm$11.52&10.77$\pm$5.86&3.42$\pm$1.95\\
5&0.95 &0.90  & 0.70&82.22$\pm$9.93&73.31$\pm$11.72&\textbf{7.69$\pm$3.75}&\textbf{2.60$\pm$1.40}\\ 
6&0.95 &0.90  & 0.50&82.17$\pm$10.99&73.12$\pm$12.42&8.88$\pm$4.53&3.10$\pm$1.77\\
\noalign{\smallskip} 
\hline
\noalign{\smallskip}
7&- &0.50  & 0.50&80.82$\pm$11.60&71.22$\pm$13.26&11.17$\pm$6.85&3.58$\pm$2.11\\ 

\noalign{\smallskip} 
\hline
\noalign{\smallskip}

\end{tabular}}
\end{threeparttable}
\end{table}
Both \{1,2,3\} and \{4,5,6\} show that $t_s=0.7$ obtain the best performance. The result complies with our previous analysis. When $t_s$ is too high, correctly predicted voxels in reliable prediction are wasted. And when $t_s$ is low, predictions with extreme low confidence are selected as pseudo label, which introduces much noise to the cross-teaching. Setting $t_q=0.98$ performs better than setting $t_q=0.95$, and it indicates the criterion of selecting reliable prediction cannot be too loose. The result of hyper-parameters set 7 shows that when we set hard and soft thresholds equally low, the performance is largely degraded, and it validates the effectiveness of our hard-soft threshold strategy.

\section{Conclusion}
In this paper, we extend sparse annotation to cross annotation to suit more general real clinical scenario. We label slices from two planes and it enlarges the diversity of annotation. To better leverage the cross annotation, we view the problem from the perspective of semi-supervised segmentation and we propose a novel cross-teaching paradigm which imposes consistency on the prediction of 3D and 2D networks. Furthermore, to achieve robust cross-supervision, we propose new strategies to select credible pseudo label, which are hard-soft threshold for 3D network and consistent prediction fusion for 2D networks. And the result on MMWHS dataset validates the effectiveness of our method.

\subsubsection{Acknowledgements} This work is supported by the Science and Technology Innovation 2030 New Generation Artificial Intelligence Major Projects (2021ZD011\\3303), NSFC Program (62222604, 62206052, 62192783), China Postdoctoral Science Foundation Project (2021M690609), Jiangsu NSF Project (BK20210224), Shandong NSF (ZR2023MF037) and CCF-Lenovo Bule Ocean Research Fund.

 \bibliographystyle{splncs04}
 \bibliography{ref}

\end{document}